\pdfoutput=1
\documentclass[11pt]{article}

\usepackage{EMNLP2022}

\usepackage{times}
\usepackage{latexsym}

\usepackage[T1]{fontenc}
\usepackage[utf8]{inputenc}

\usepackage{microtype}

\usepackage{graphicx}
\graphicspath{ {./images/} }

\usepackage{inconsolata}
\usepackage{caption}
\usepackage{subcaption}

\usepackage{grffile}

\usepackage{hyperref}

\usepackage{natbib}

\title{ClassBases at CASE-2022 Multilingual Protest Event Detection Tasks:\\ Multilingual Protest News Detection and \\ Automatically Replicating Manually Created Event Datasets}

\author{Peratham Wiriyathammabhum \\
  \texttt{peratham.bkk@gmail.com} \\
  \vspace{-4\baselineskip}
  }

\begin{document}
\maketitle
\begin{abstract}
In this report, we describe our ClassBases submissions to a shared task on multilingual protest event detection. For the multilingual protest news detection, we participated in subtask-1, subtask-2, and subtask-4, which are document classification, sentence classification, and token classification. In subtask-1, we compare XLM-RoBERTa-base, mLUKE-base, and XLM-RoBERTa-large on finetuning in a sequential classification setting. We always use a combination of the training data from every language provided to train our multilingual models. We found that larger models seem to work better and entity knowledge helps but at a non-negligible cost. For subtask-2, we only submitted an mLUKE-base system for sentence classification. For subtask-4, we only submitted an XLM-RoBERTa-base for token classification system for sequence labeling. For automatically replicating manually created event datasets, we participated in COVID-related protest events from the New York Times news corpus. We created a system to process the crawled data into a dataset of protest events. 
\end{abstract}

\begin{figure*}[t]
\centering
   \begin{subfigure}[b]{0.23\textwidth}
         \centering
         \includegraphics[width=\textwidth]{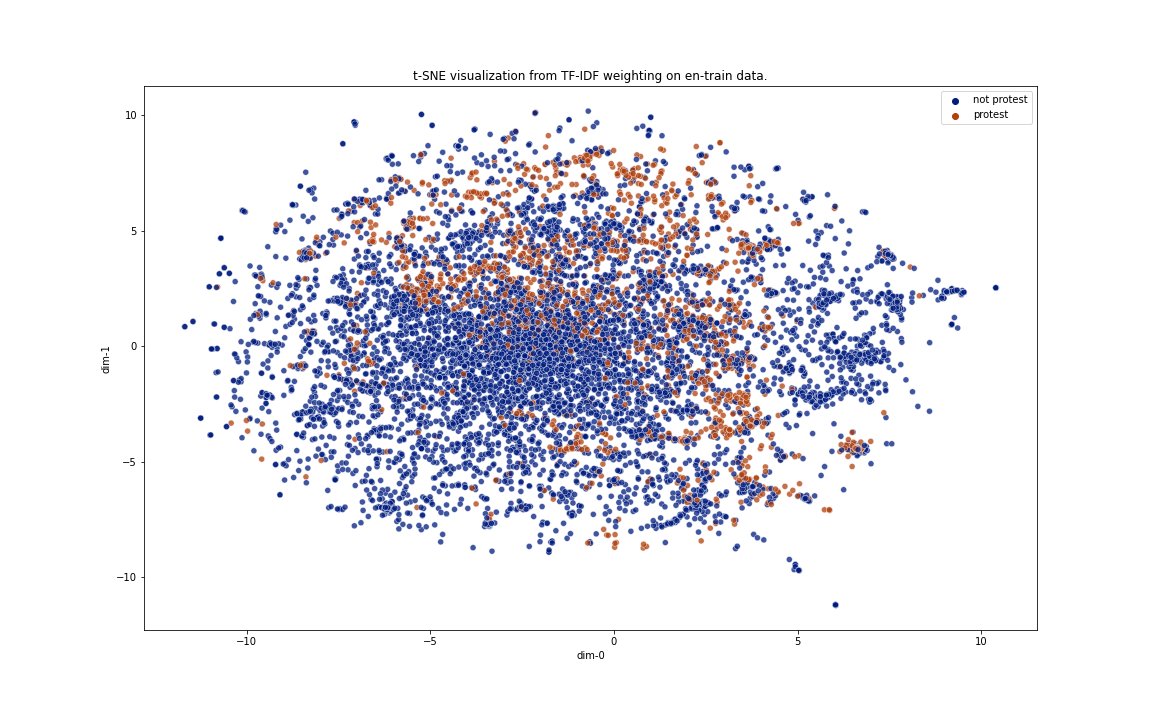}
         \caption{English (en)}
         \label{fig:en}
     \end{subfigure}
     \hfill
    \begin{subfigure}[b]{0.23\textwidth}
         \centering
         \includegraphics[width=\textwidth]{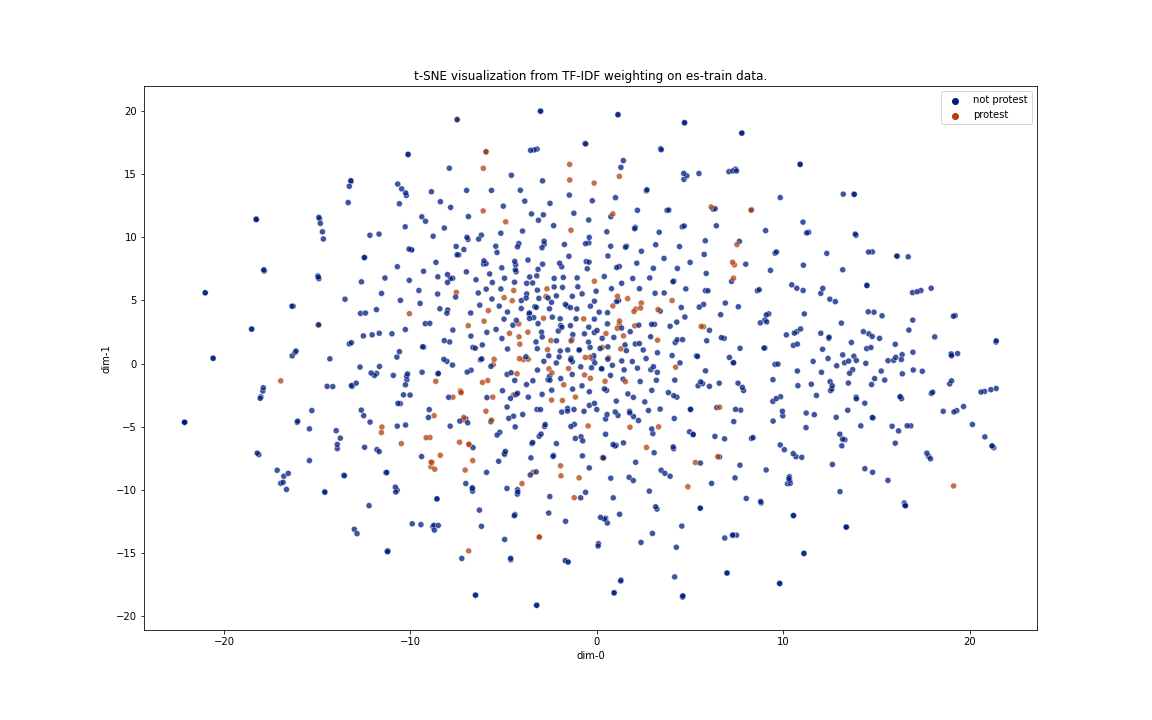}
         \caption{Spanish (es)}
         \label{fig:es}
     \end{subfigure}
     \hfill
    \begin{subfigure}[b]{0.23\textwidth}
         \centering
         \includegraphics[width=\textwidth]{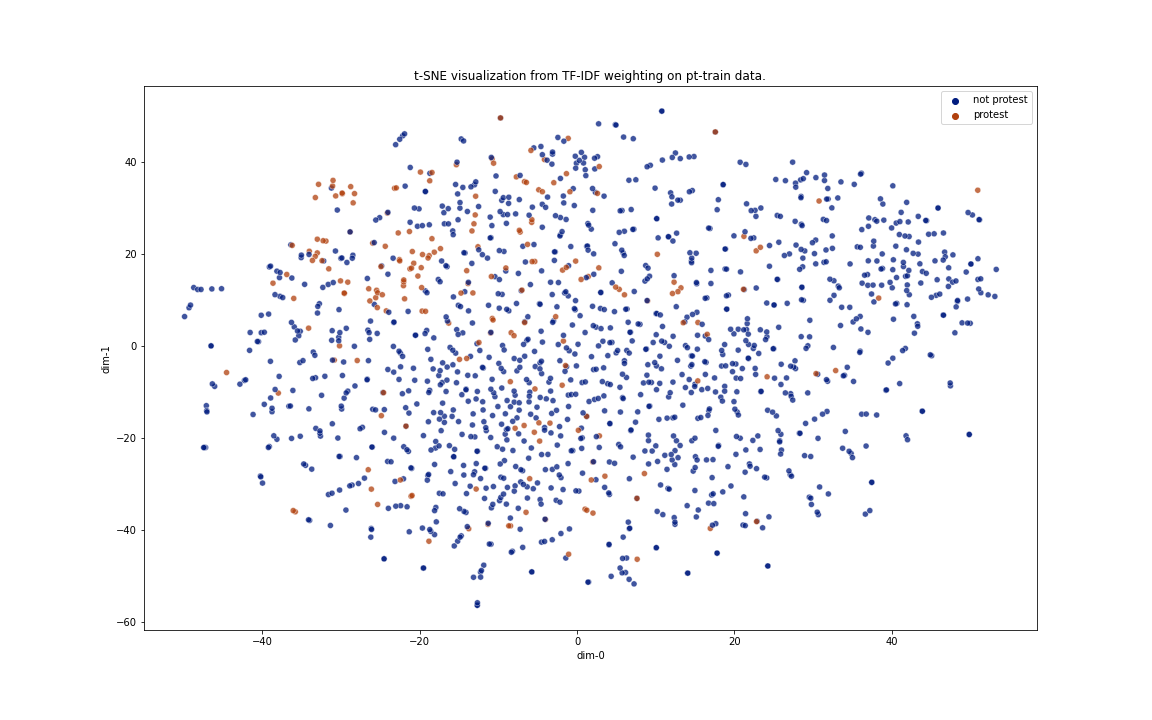}
         \caption{Portuguese (pt)}
         \label{fig:pt}
     \end{subfigure}
     \hfill
     \begin{subfigure}[b]{0.23\textwidth}
         \centering
         \includegraphics[width=\textwidth]{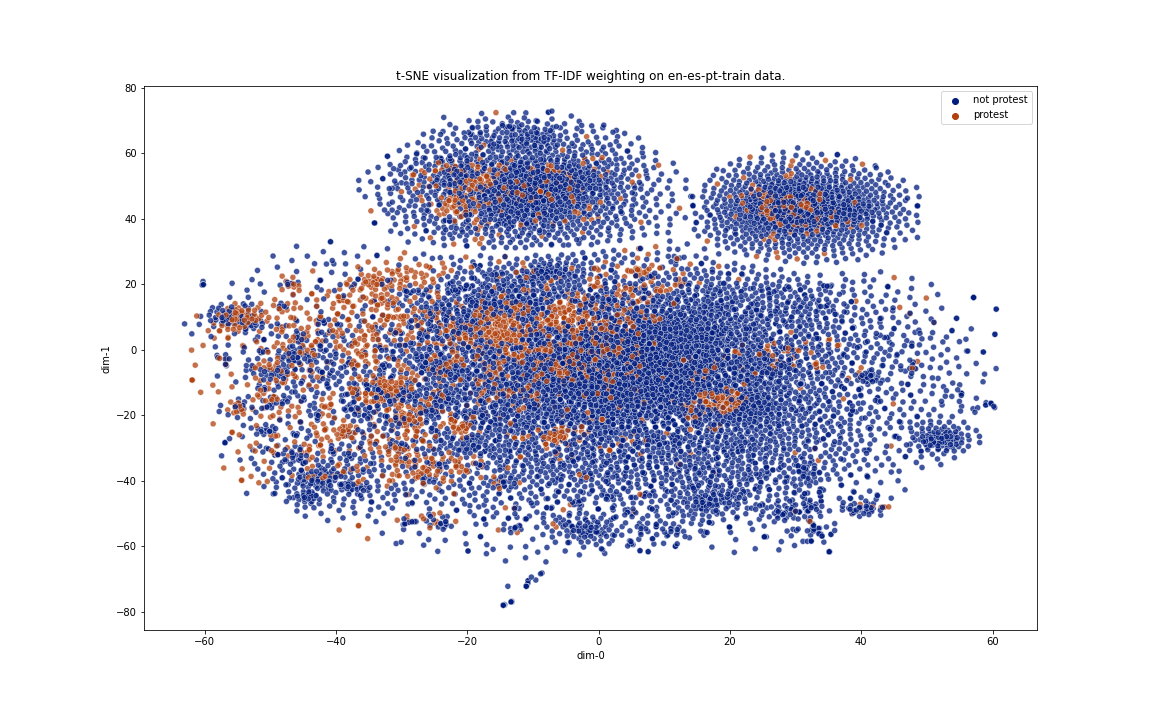}
        \caption{en-es-pt concatenated}
         \label{fig:concat}
     \end{subfigure}
% \vspace{-1.0\baselineskip}
   \caption{\textbf{The distribution of tf-idf weighted subtask1 training set document data visualized using t-SNE \cite{van2008visualizing}. The blue dots have no protest event, and the orange dots have some protest events.}}
% \label{fig:long}
% \label{fig:onecol}
\label{figplot}
\vspace{-0.5\baselineskip}
\end{figure*}

\begin{figure*}[t]
\centering
\begin{subfigure}[b]{0.23\textwidth}
         \centering
         \includegraphics[width=\textwidth]{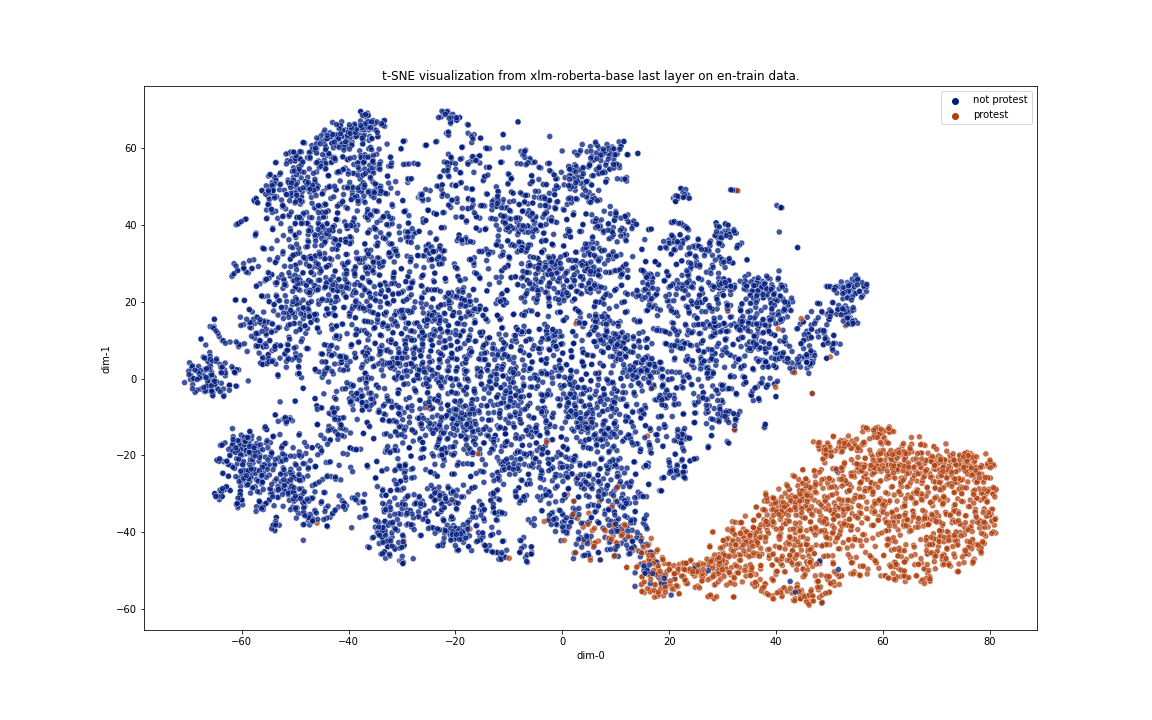}
         \caption{English (en)}
         \label{fig:enxlmrbase}
     \end{subfigure}
     \hfill
    \begin{subfigure}[b]{0.23\textwidth}
         \centering
         \includegraphics[width=\textwidth]{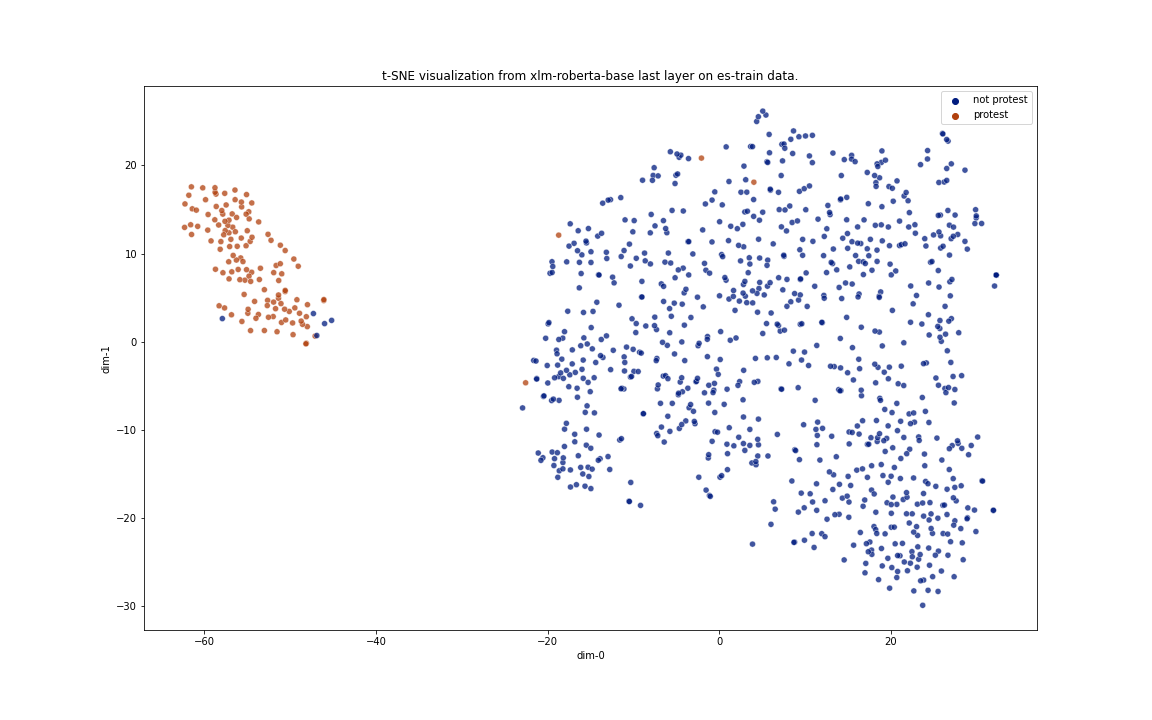}
         \caption{Spanish (es)}
         \label{fig:esmxlmrbase}
     \end{subfigure}
     \hfill
    \begin{subfigure}[b]{0.23\textwidth}
         \centering
         \includegraphics[width=\textwidth]{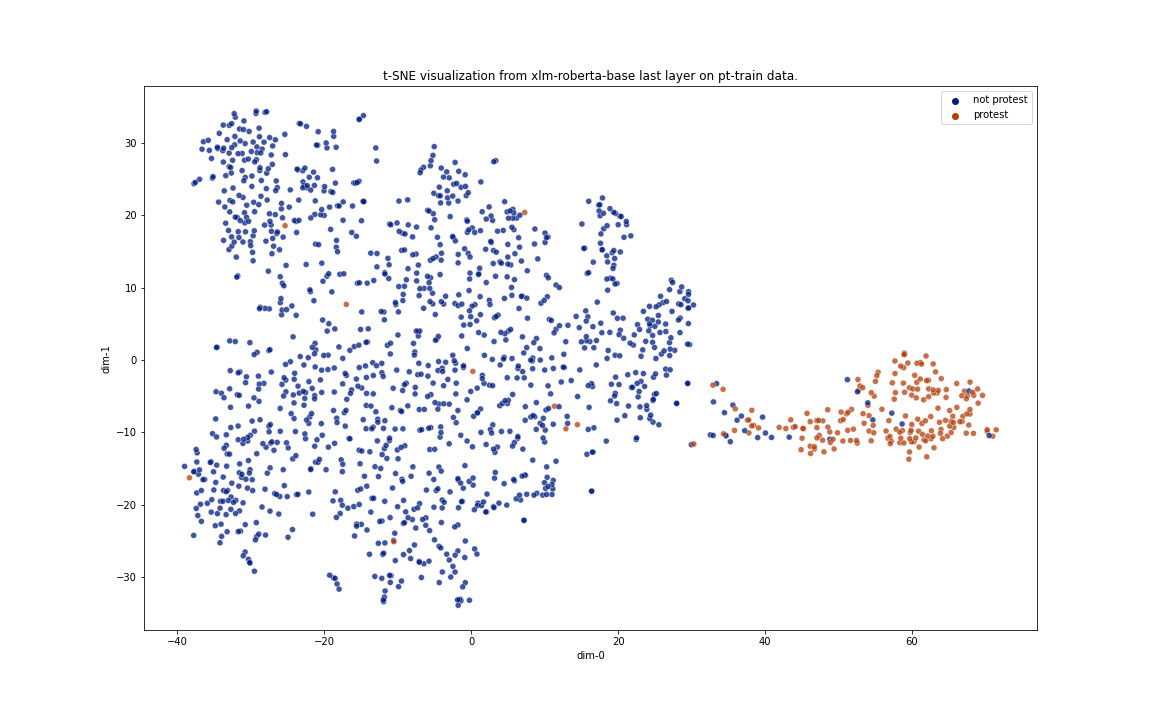}
         \caption{Portuguese (pt)}
         \label{fig:ptmxlmrbase}
     \end{subfigure}
     \hfill
     \begin{subfigure}[b]{0.23\textwidth}
         \centering
         \includegraphics[width=\textwidth]{images/sns_en-xlm-roberta-base-last-tsne.png}
         \caption{en-es-pt concatenated}
         \label{fig:concatxlmrbase}
     \end{subfigure}
     \hfill
   \begin{subfigure}[b]{0.23\textwidth}
         \centering
         \includegraphics[width=\textwidth]{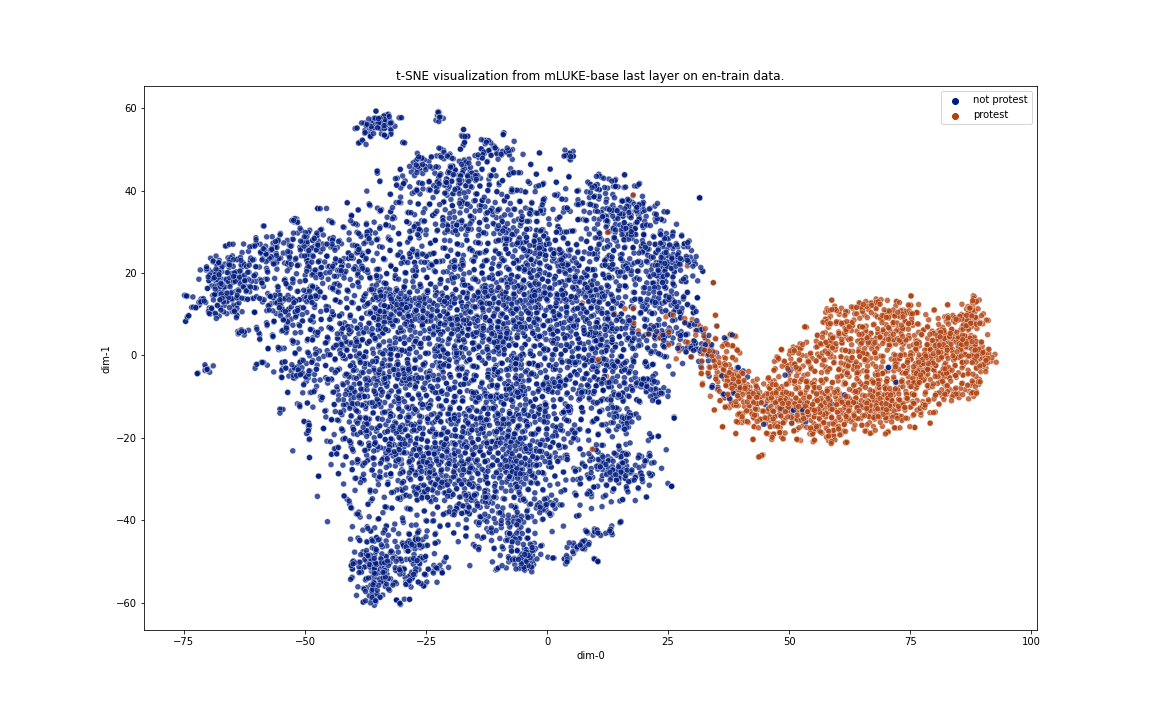}
         \caption{English (en)}
         \label{fig:enmluke}
     \end{subfigure}
     \hfill
    \begin{subfigure}[b]{0.23\textwidth}
         \centering
         \includegraphics[width=\textwidth]{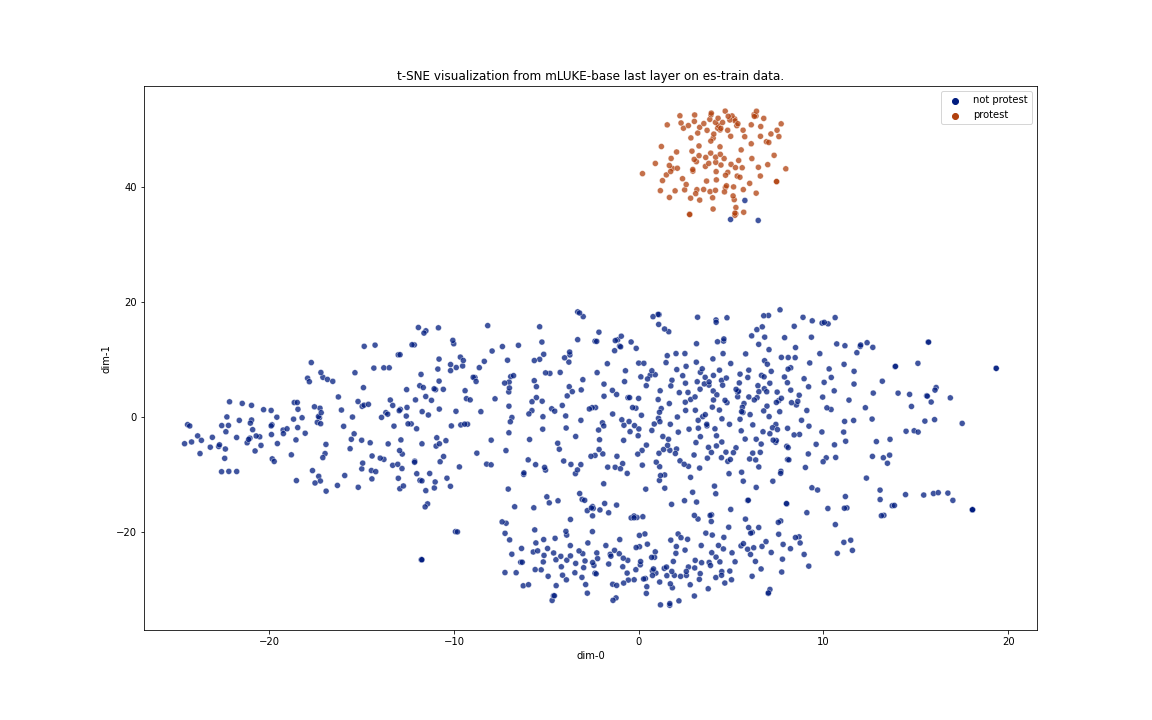}
         \caption{Spanish (es)}
         \label{fig:esmluke}
     \end{subfigure}
     \hfill
    \begin{subfigure}[b]{0.23\textwidth}
         \centering
         \includegraphics[width=\textwidth]{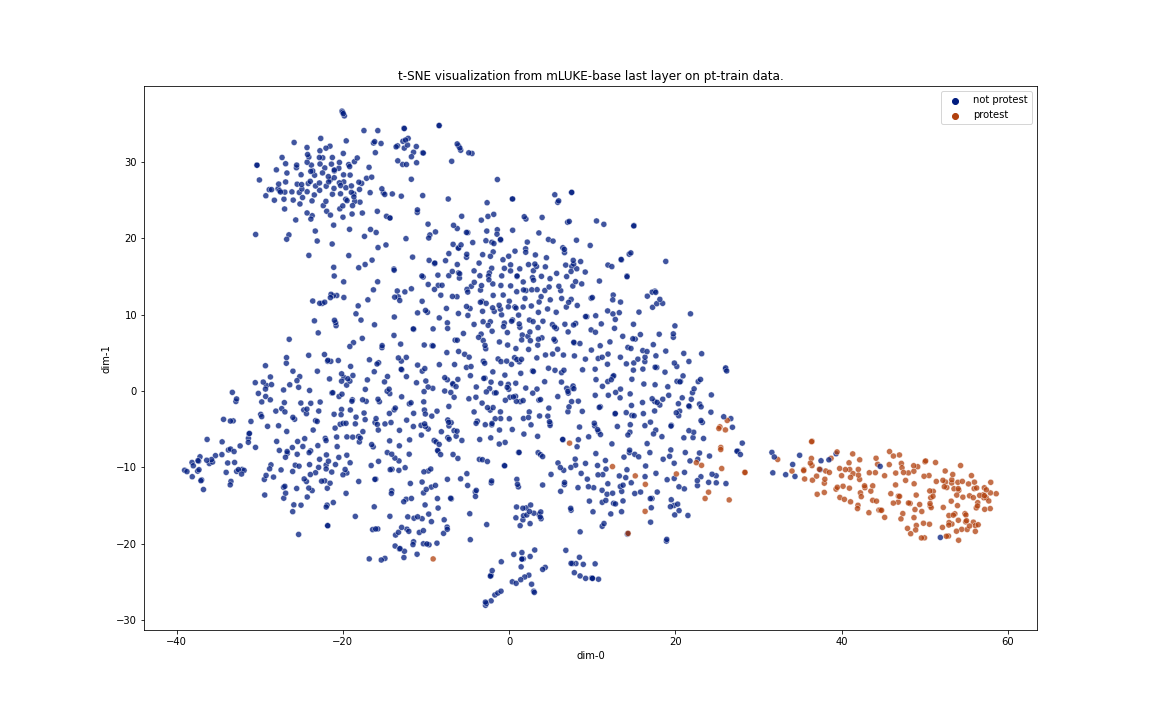}
         \caption{Portuguese (pt)}
         \label{fig:ptmluke}
     \end{subfigure}
     \hfill
     \begin{subfigure}[b]{0.23\textwidth}
         \centering
         \includegraphics[width=\textwidth]{images/sns_en-mLUKE-base-last-tsne.png}
         \caption{en-es-pt concatenated}
         \label{fig:concatmluke}
     \end{subfigure}
     \hfill
     \begin{subfigure}[b]{0.23\textwidth}
         \centering
         \includegraphics[width=\textwidth]{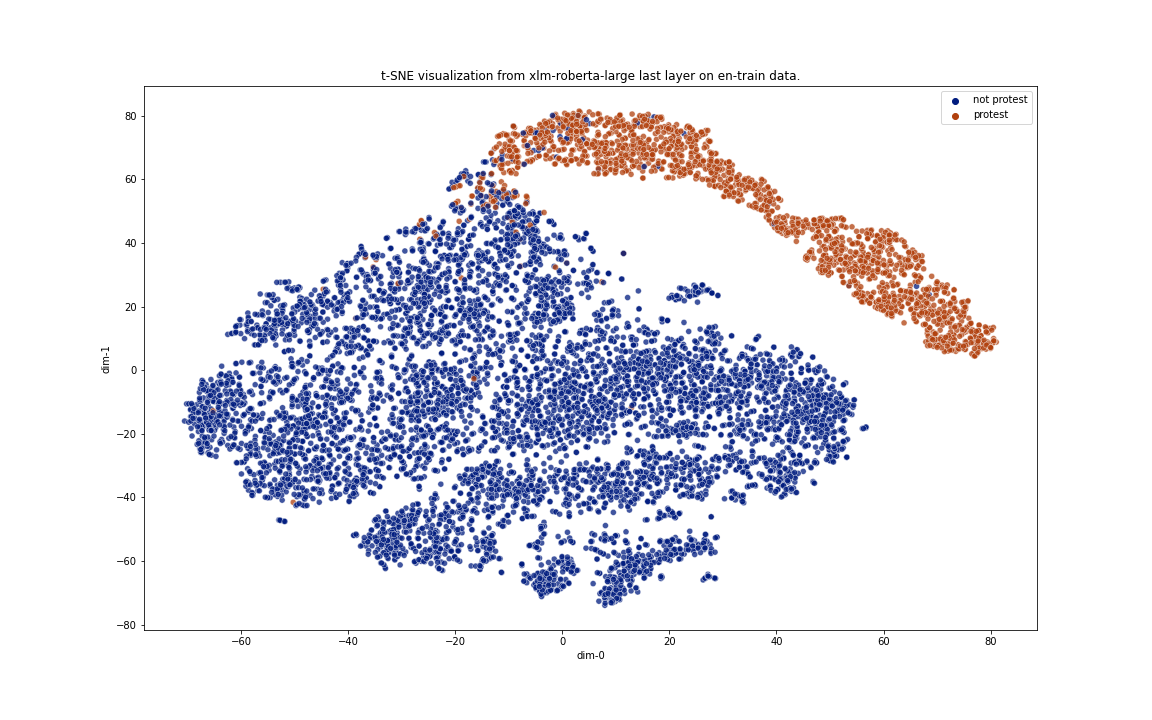}
         \caption{English (en)}
         \label{fig:en2}
     \end{subfigure}
     \hfill
    \begin{subfigure}[b]{0.23\textwidth}
         \centering
         \includegraphics[width=\textwidth]{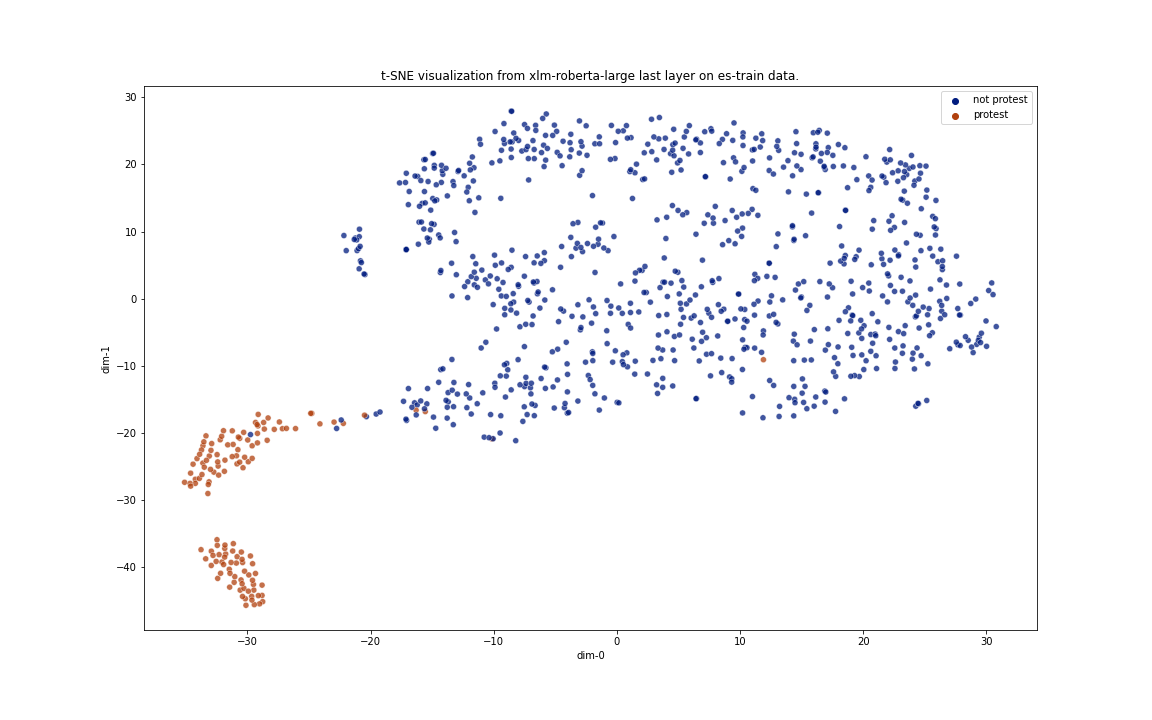}
         \caption{Spanish (es)}
         \label{fig:es2}
     \end{subfigure}
     \hfill
    \begin{subfigure}[b]{0.23\textwidth}
         \centering
         \includegraphics[width=\textwidth]{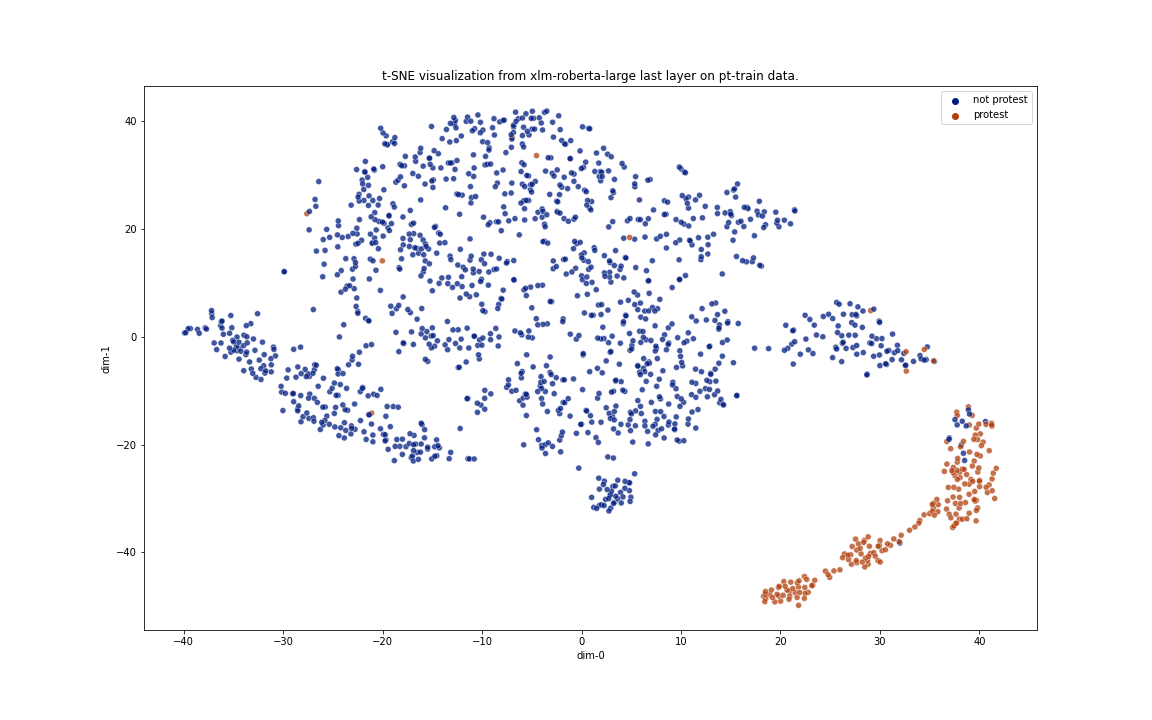}
         \caption{Portuguese (pt)}
         \label{fig:pt2}
     \end{subfigure}
     \hfill
     \begin{subfigure}[b]{0.23\textwidth}
         \centering
         \includegraphics[width=\textwidth]{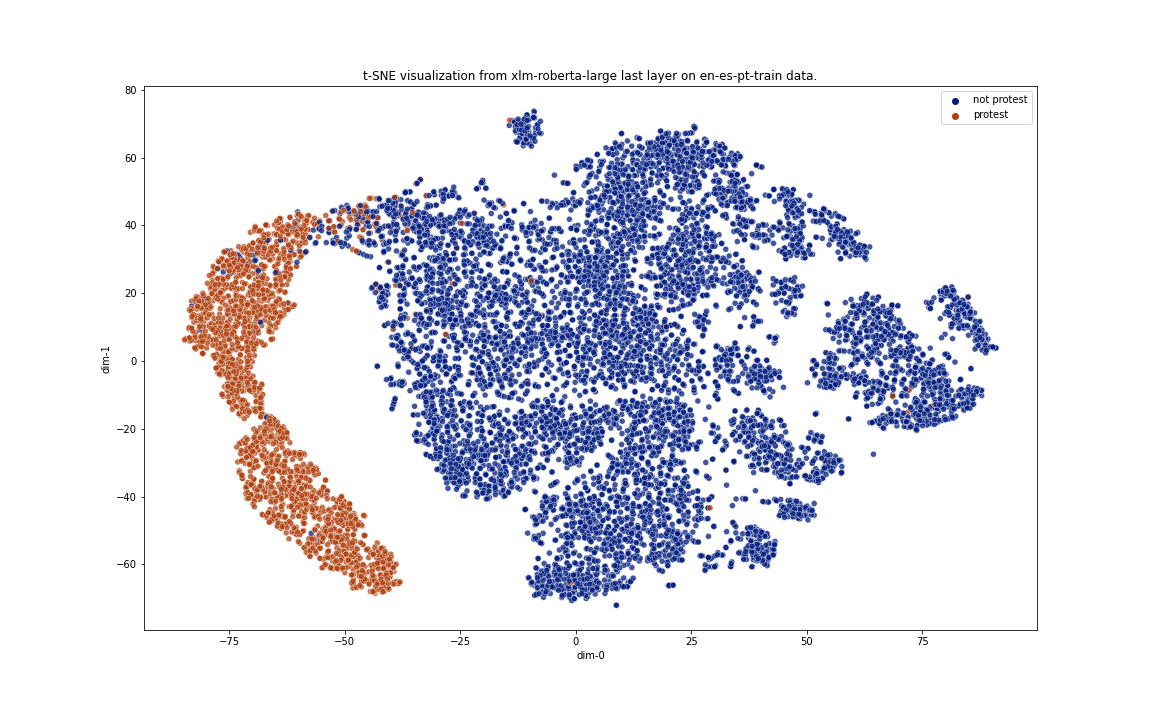}
         \caption{en-es-pt concatenated}
         \label{fig:concat2}
     \end{subfigure}
% \vspace{-1.0\baselineskip}
   \caption{\textbf{The distribution of subtask1 training set document features extracted by averaging over the sequence dimension of the last layer from our finetuned XLM-RoBERTa-base (the first row), mLUKE-base (the second row), and XLM-RoBERTa-large (the third row) visualized using t-SNE \cite{van2008visualizing}. The blue dots have no protest event, and the orange dots have some protest events.}}
% \label{fig:long}
% \label{fig:onecol}
\label{figplot2}
\vspace{-0.5\baselineskip}
\end{figure*}

% \begin{figure*}[t]
% \centering
%   \begin{subfigure}[b]{0.23\textwidth}
%          \centering
%          \includegraphics[width=\textwidth]{images/sns_en-xlm-roberta-large-last-tsne.png}
%          \caption{English (en)}
%          \label{fig:en2}
%      \end{subfigure}
%      \hfill
%     \begin{subfigure}[b]{0.23\textwidth}
%          \centering
%          \includegraphics[width=\textwidth]{images/sns_es-xlm-roberta-large-last-tsne.png}
%          \caption{Spanish (es)}
%          \label{fig:es2}
%      \end{subfigure}
%      \hfill
%     \begin{subfigure}[b]{0.23\textwidth}
%          \centering
%          \includegraphics[width=\textwidth]{images/sns_pt-xlm-roberta-large-last-tsne.png}
%          \caption{Portuguese (pt)}
%          \label{fig:pt2}
%      \end{subfigure}
%      \hfill
%      \begin{subfigure}[b]{0.23\textwidth}
%          \centering
%          \includegraphics[width=\textwidth]{images/sns_en-es-pt-xlm-roberta-large-last-tsne.png}
%          \caption{en-es-pt concatenated}
%          \label{fig:concat2}
%      \end{subfigure}
% % \vspace{-1.0\baselineskip}
%   \caption{\textbf{The distribution of subtask1 training set document features extracted by averaging over the sequence dimension of the last layer from our finetuned XLM-RoBERTa-large visualized using t-SNE \cite{van2008visualizing}. The blue dots have no protest event, and the orange dots have some protest events.}}
% % \label{fig:long}
% % \label{fig:onecol}
% \label{figplot2}
% \vspace{-0.5\baselineskip}
% \end{figure*}

\section{Introduction}
A shared task on multilingual protest event detection at CASE-2022 is the second installment from the previous event at CASE-2021 about socio-political and crisis events detection \cite{hurriyetoglu-etal-2021-multilingual, 10.1162/dint_a_00092}. The shared task focuses on protest events where people complain, put their objections, or display their unwillingness to a course of action whether that action is from an authority or a government \cite{mw:protest}. 

As in the previous installment, this shared task organizes the automated multilingual protest event detection pipeline into multiple subsequent steps at different granularity levels as four subtasks, document classification, sentence classification, event sentence coreference identification, and event extraction. Moreover, the shared task contains many languages in many different magnitudes of data sizes, from ten thousand data points to hundreds of data points to no data points. In other words, many settings are varying from full training to low-resource training to few-shot learning to zero-shot learning. 

\begin{itemize}
  \vspace{-0.3\baselineskip}
\item The first subtask, \textit{document classification}, tries to classify whether a given document, a piece of news, or an article, contains any information about a past or an ongoing socio-political protest event. The shared task provides a full training setting for English, Spanish and Portuguese on a scale of thousands of data points. Then, there is a zero-shot training setting for Hindi, Turkish, Urdu, and Mandarin. 
  \vspace{-0.3\baselineskip}
\item The second subtask, \textit{sentence classification}, classifies whether a given sentence from a document contains any information about a past or an ongoing socio-political protest event. The shared task provides a full training setting for English, Spanish and Portuguese on the scale of ten thousand data points for English and thousands of data points for Spanish and Portuguese.
  \vspace{-0.3\baselineskip}
\item The third subtask, \textit{event sentence coreference identification}, tries to group sentences, from the same document, containing socio-political events from the same stories together. There are hundreds of training instances for English and around twenty training instances for Spanish and Portuguese. 
  \vspace{-0.3\baselineskip}
\item The fourth subtask, \textit{event extraction}, extracts event entity spans, triggers, and arguments, from event sentences within the same story. 
  \vspace{-0.3\baselineskip}
\end{itemize}

We participate in the first, second, and fourth subtasks. We build our system solutions upon Huggingface's multilingual transformer language models \cite{wolf-etal-2020-transformers}, specifically, XLM-RoBERTa language models \cite{conneau-etal-2020-unsupervised} and mLUKE multilingual transformer language models with entity embedding \cite{ri-etal-2022-mluke}. We also participated in creating COVID-related protest event datasets from the New York Times news corpus \cite{Zavarella22a}. The codes for our systems are open-sourced and available at our GitHub repository\footnote{\url{https://github.com/perathambkk/case_shared_task_emnlp2022}}.

\section{Models}
As in the IBM MNLP team report \cite{awasthy-etal-2021-ibm}, whose systems top-scored in most subtasks of the previous CASE-2021, we consider XLM-RoBERTa language models (XLM-R) \cite{conneau-etal-2020-unsupervised} trained on the concatenation of the data from all languages available from the shared task. XLM-RoBERTa built upon the RoBERTa language model \cite{liu2019roberta} and multilingual pretrained on 2.5 TB of filtered CommonCrawl data consisting of 100 languages. By pretraining jointly across many multiple languages, hopefully, the model can transfer information across languages. However, the paper indicates the \textit{curse of multilinguality} trade-off where we can scale the number of languages up to the point that the model performance for low-resource languages starts to degrade. Still, XLM-RoBERTa seems not to suffer from this trade-off yet by increasing the model capacity and performing very well on many benchmarks. 

We also consider mLUKE, a multilingual transformer language model with entity embeddings, \cite{ri-etal-2022-mluke}. The mLUKE language model is also based on XLM-RoBERTa but has an optional entity embedding set for downstream tasks and was pretrained on 24 languages using Wikipedia. The entity embeddings are cross-lingual mappings of entities learned from Wikipedia. The language model part was pretrained as a masked language model and the entity embedding part was pretrained in a masked entity prediction task. Despite the performance gains on entity-related downstream tasks, a major limitation of incorporating entity embeddings is the large memory footprint. That is, using only an mLUKE-base model requires about the same GPU memory as an XLM-RoBERTa-large model. 

\begin{figure}[t]
\begin{center}
   \includegraphics[width=1.00\linewidth]{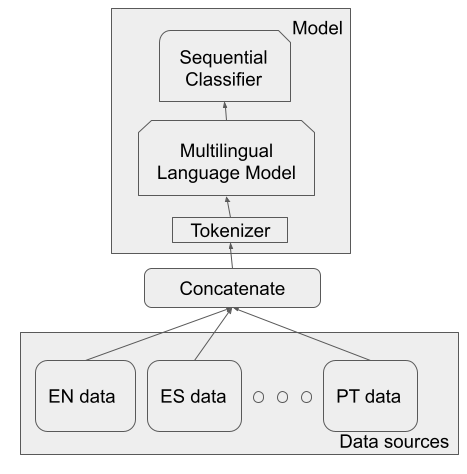}
\end{center}
\vspace{-1.0\baselineskip}
   \caption{\textbf{The architecture of our systems. We concatenated data from all languages and randomly sample them into batches. The batches are inputs to our models. The model part consists of a tokenizer, a multilingual language model, and a sequential classifier, all are from the Huggingface's library \cite{wolf-etal-2020-transformers}. For subtask4, we replace a sequential classifier with a token classifier.}}
\label{fig:long}
\label{fig:onecol}
\label{figarch}
\vspace{-1\baselineskip}
\end{figure}

\section{Experimental Results}
All of our experiments were done in the Google Colab setting on NVIDIA Tesla T4 GPUs. We used the batch size in the range of $16-36$ and the learning rate for an AdamW optimizer \cite{loshchilov2018decoupled} in the set of $\{2.5e-5, 5e-5\}$ for all experiments. We considered a linear annealing scheduler. Also, adding a warm-up step does not make any difference so we set the warm-up step to zero in all experiments. 

Except otherwise stated, we concatenated the given training data in all languages as our combined training set for every subtask. We also employed the early stopping with zero patience training strategy schema \cite{prechelt1998early, bengio2012practical}. We varied the training epoch until the training metric saturated with manual monitoring, and then stopped right at the end of that epoch. However, we mostly tried with one or two candidate numbers of training epochs since training large language models takes a few hours and Gooogle's Colab GPU time just runs out.

\begin{table*}[t] 
\centering
\caption{Test macro F1-scores of our models in subtask1: Document Classification 2021 test data. (The numbers in subscript are submission rankings on the leaderboard from our best submissions. The symbol $\dagger$ denotes the result is better than the previous CASE-21 best submission.)}
\vspace{-0.5\baselineskip}
\label{st1table}
\begin{tabular}{|l|c|c|c|c|}
\hline
 Model    & en & pt & es & hi \\
\hline
XLM-R-base  & $79.82$ & $79.55$ & $68.70$ & $79.35$\\
mLUKE-base  & $79.91$ & $80.02$ & $72.93$ & $75.77$\\
XLM-R-large  & $\mathbf{82.30_4}$ &$\mathbf{85.39_2}\dagger$ & $\mathbf{73.48_4}$ & $\mathbf{80.77_1}\dagger$\\
\hline
CASE-21 best & 84.55& 84.00 & 77.27& 78.77\\
\cite{hurriyetoglu-etal-2021-multilingual} &&&&\\
\hline
\end{tabular}
% \vspace{-1\baselineskip}
\end{table*}

\begin{table*}[t] 
\centering
\caption{Test macro F1-scores of our models in subtask1: Document Classification 2021+2022 test data. (The numbers in subscript are submission rankings on the leaderboard from our best submissions.)}
\vspace{-0.5\baselineskip}
\label{st1tablec}
\begin{tabular}{|l|c|c|c|c|c|c|c|}
\hline
 Model    & en & pt & es & hi & tr & ur & zh\\
\hline
mLUKE-base  & $77.35$ & $74.67$ & $\mathbf{69.25_6}$ & $69.54$ & $\mathbf{78.57_5}$ & $67.91$ & $73.79$\\
XLM-R-large  & $\mathbf{78.50_6}$ &$\mathbf{77.11_5}$ & $66.86$ & $\mathbf{80.78_1}$ & $75.66$ & $\mathbf{75.72_5}$ & $\mathbf{77.16_5}$\\
\hline
\end{tabular}
% \vspace{-1\baselineskip}
\end{table*}

\subsection{Document Classification}
We trained XLM-RoBERTa-base, XLM-RoBERTa-large, and mLUKE-base as sequence classifiers for document classification. The models classify whether a given document contains any protest events or not as a binary classification task. The input document is truncated to the maximum length of $150$. Then, the truncated document is fed into a transformer language model with a softmax layer on top which outputs logits for binary classifications. We trained XLM-RoBERTa-base for $12$ epochs, mLUKE-base for $15$ epochs, and XLM-RoBERTa-large for $5$ epochs, respectively. We used the batch size of $36$ for base models, XLM-RoBERTa-base and mLUKE-base, and we used the batch size of $16$ for our large model, XLM-RoBERTa-large.

The experimental results in Table \ref{st1table} suggest that a small model (XLM-RoBERTa-base) does not perform well in general. However, adding entity knowledge makes a small model (mLUKE-base) performs much better typically at a cost except in Hindi where mLUKE-base might be trained on less number of languages and does not perform well in the zero-shot setting. Still, a larger language model (XLM-RoBERTa-large) performs best most of the time. Surprisingly, our XLM-RoBERTa-large submissions perform better than the best submissions from the previous year in Portuguese and Hindi using only a single model and without any external data. In the previous CASE-21, the best Portuguese submission uses an ensemble and the best Hindi submission uses some external data so it is not a zero-shot setting. 

We visualized the tf-idf weighted training data in Figure \ref{figplot} using t-SNE \cite{van2008visualizing, wattenberg2016how}. The scatter plots show the inseparability of the class data, and the concatenated data plot in Figure \ref{figplot}(\subref{fig:concat}) shows that the data in each language are in different regions. However, the visualization of the XLM-RoBERTa-base, mLUKE-base, and XLM-RoBERTa-large features shows that the finetuned multilingual language models cram the data from various languages into the same space by their class information. The plots in the same row from Figure \ref{figplot2} are all the same shapes. 

This year, the shared task organizers provide a new test set that contains more data and more languages \cite{Hurriyetoglu+22a}. There are Turkish, Urdu, and Mandarin test data in addition to the existing English, Portuguese, Spanish, and Hindi. We also tested our models in this setting where Hindi, Turkish, Urdu, and Mandarin were tested in zero-shot settings. We compare mLUKE-base and XLM-RoBERTa-large in Table \ref{st1tablec}. From the results, mLUKE-base works better in Spanish and Turkish while XLM-RoBERTa-large works best for the remaining languages. The results are not consistent for zero-shot setting languages, however, XLM-RoBERTa-large works better $3$ out of $4$ cases. Also, in the low-resource settings, mLUKE-base works better in Spanish while XLM-RoBERTa-large works better in Portuguese. 

\subsection{Sentence Classification}
We trained XLM-RoBERTa-large and mLUKE-base as sequence classifiers for sentence classification. Similar to document classification, we set the maximum sentence length to $150$ and fed a sentence to a transformer language model with a softmax layer on top. In this subtask, we trained each model for $2.30$ hours. We trained mLUKE-base for $15$ epochs with a batch size of $36$ and XLM-RoBERTa-large for $6$ epochs with a batch size of $30$ (a batch size of $10$ with a gradient accumulation step of $3$.). We observed that mLUKE-base was converged but XLM-RoBERTa-large was just fitted to a degree given the same resource. 

The experimental results in Table \ref{st2table} suggest that mLUKE-base works better in low-resource languages, Portuguese and Spanish, while XLM-RoBERTa-large works better in English despite being undertrained. Our submissions are not better than the previous year's best results in this subtask. 

\begin{table}[!t] 
\centering
\caption{Test macro F1-scores of our models in subtask1: Sentence Classification 2021 test data. (The numbers in subscript are submission rankings on the leaderboard from our best submissions. The best results from the previous year are from \cite{hurriyetoglu-etal-2021-multilingual}.)}
\vspace{-0.5\baselineskip}
\label{st2table}
\begin{tabular}{|l|c|c|c|}
\hline
 Model    & en & pt & es  \\
\hline
mLUKE-base  & $79.65$ & $\mathbf{86.83_3}$ & $\mathbf{87.10_4}$ \\
XLM-R-large  & $\mathbf{81.12_4}$ &$85.39$ & $84.62$ \\
\hline
CASE-21 best & 85.32& 88.47&88.61 \\
\hline
\end{tabular}
% \vspace{-1\baselineskip}
\end{table}
% \subsection{Event Sentence Coreference Identification}
% This is sad to say, we did not participate in this subtask.

\subsection{Event Extraction}
We only trained an XLM-RoBERTa-base model for token classification. We split the data into training and validation using the ratio of $0.2$. However, there are so few Portuguese and Spanish data and XLM-RoBERTa-base does not have enough capacity so it does not perform well in our experiments as shown in Table \ref{st4table}, sadly. 

We speculate that some training strategy, which does not require data partitioning, and larger language models will perform better in this subtask. 

\begin{table}[!t] 
\centering
\caption{Test CoNLL F1-scores of our models in subtask4: Event Extraction. (The numbers in subscript are submission rankings on the leaderboard.)}
\vspace{-0.5\baselineskip}
\label{st4table}
\begin{tabular}{|l|c|c|c|}
\hline
 Model    & en & pt & es  \\
\hline
XLM-R-base  & $46.88_5$ & $12.53_5$ & $37.10_5$ \\
\hline
\end{tabular}
% \vspace{-1\baselineskip}
\end{table}

\subsection{Automatically Replicating Manually Created Event Datasets}
In this task \cite{Zavarella22a}, event detection systems are going to be evaluated on their spatio-temporal pattern extraction performance. Similar to the previous shared task installment on Black Lives Matter \cite{giorgi-etal-2021-discovering}, this year's target event is COVID-related protests in the US spanning three months (July 27, 2020 through October 27, 2020). We adopt our components from last year's report. 

To begin with, we used the trained XLM-RoBERTa-large from subtask1 to classify the news using a concatenation of its news title and news abstract to see whether it contains any protest events or not. If the classifier outputs positive (logits were thresholded at $0.9$), we ran a SpaCy named entity recognizer \cite{spacy} on the textual concatenation to get spans with location tags (`GPE'). Then, those spans were concatenated into a query string which we used a geocoder library\footnote{\href{https://geocoder.readthedocs.io/}{https://geocoder.readthedocs.io/}} to geocode using the Bing Maps REST Services API\footnote{\href{https://learn.microsoft.com/en-us/bingmaps/rest-services/}{https://learn.microsoft.com/en-us/bingmaps/rest-services/}}. We used the provided dates from the date column as outputs given the filtered ids. Finally, we created a row for each filtered id containing five tuples, which are the id, the date, the city, the region or state, and the country.
\section{Conclusions}
This report describes our systems for a shared task on multilingual protest event detection at CASE-2022. We compared a small multilingual language model (XLM-RoBERTa-base), a knowledge-based multilingual model (mLUKE-base), and a large multilingual language model (XLM-RoBERTa-large). From all experimental results, we observed consistent outperforms from XLM-RoBERTa-large over smaller language models, XLM-RoBERTa-base, and mLUKE-base. Therefore, we concluded that language model capacity matters a lot for multilingual tasks. Also, we observed that mLUKE-base mostly outperforms XLM-RoBERTa-large. Hence, incorporating entity knowledge helps improve performance but with a nonnegligible computational cost. From our visualizations, we found that our finetuned multilingual language models cram data from various languages into the same space by their class information. 

\section*{Limitations}
This report is like a class assignment, given our work progress depicted here. We only compared several multilingual language models and implemented some basic systems to solve the tasks.

The authors are self-affiliated and do not represent any entities. The authors also participated in the shared task under many severe unattended local personal criminal events in their home countries. There might be some unintentional errors and physical limitations based on these unlawful interruptions. Even at the times of drafting this report, the authors suffer from unknown toxin flumes spraying into their places. We want to participate in the shared task because it is fun and educational. We apologize for any errors in this report. We tried our best.

\section*{Ethics Statement}
Scientific work published at EMNLP 2022 must comply with the \href{https://www.aclweb.org/portal/content/acl-code-ethics}{ACL Ethics Policy}. We, the authors, intend the uses of our systems for peace and social good only. No harm. To see and alleviate people dangers, pains, and angers, detecting these socio-political and crisis events is meant to be helpful and savior for all, not the other way around.

\section*{Acknowledgments}
We would like to thank the reviewers for their constructive feedback. 

% Entries for the entire Anthology, followed by custom entries
\bibliography{anthology,custom}
\bibliographystyle{acl_natbib}

\end{document}